\RequirePackage{fix-cm}
\documentclass[smallextended]{svjour3}       % onecolumn (second format)
\smartqed  % flush right qed marks, e.g. at end of proof
\usepackage{graphicx}
\usepackage[misc]{ifsym}
\begin{document}
	
	\title{Learning Transformer Features for Image Quality Assessment%\thanks{Grants or other notes
		%about the article that should go on the front page should be
		%placed here. General acknowledgments should be placed at the end of the article.}
	}
	%\subtitle{Do you have a subtitle?\\ If so, write it here}
	
	%\titlerunning{Short form of title}        % if too long for running head
	
	\author{{Chao Zeng}   \textsuperscript{1}      \and
		{Sam Kwong}   \textsuperscript{1,~\Letter} 
	}
	
	%\authorrunning{Short form of author list} % if too long for running head
	
	\institute{%
		\begin{itemize}
			\item[\textsuperscript{\Letter}] 
			{Sam Kwong}\\
			%Tel.: +123-45-678910\\
			%Fax: +123-45-678910\\
			\email{cssamk@cityu.edu.hk}
			\item[\textsuperscript{}] 
			{Chao Zeng}\\
			%Tel.: +123-45-678910\\
			%Fax: +123-45-678910\\
			\email{chao.zeng@my.cityu.edu.hk}
			\at
			\item[\textsuperscript{1}] Department of Computer Science, City University of Hong Kong, Kowloon, Hong Kong \\
		\end{itemize}
		%	C. Zeng \at
		%              City University of Hong Kong \\
		%              %Tel.: +123-45-678910\\
		%              %Fax: +123-45-678910\\
		%              \email{chao.zeng@my.cityu.edu.hk}           %  \\
		%%             \emph{Present address:} of F. Author  %  if needed
		%           \and
		%           S. Kwong \at
		%              City University of Hong Kong \\
		%              \email{cssamk@cityu.edu.hk} 
	}
	
	\date{Received: date / Accepted: date}
	% The correct dates will be entered by the editor
	
	\maketitle
	
	\begin{abstract}
		Objective image quality evaluation is a challenging task, which aims to measure the quality of a given image automatically. According to the availability of the reference images, there are Full-Reference and No-Reference IQA tasks, respectively. Most deep learning approaches use regression from deep features extracted by Convolutional Neural Networks. For the FR task, another option is conducting a statistical comparison on deep features. For all these methods, non-local information is usually neglected. In addition, the relationship between FR and NR tasks is less explored. Motivated by the recent success of transformers in modeling contextual information, we propose a unified IQA framework that utilizes CNN backbone and transformer encoder to extract features. The proposed framework is compatible with both FR and NR modes and allows for a joint training scheme. Evaluation experiments on three standard IQA datasets, i.e., LIVE, CSIQ and TID2013, and KONIQ-10K, show that our proposed model can achieve state-of-the-art FR performance. In addition, comparable NR performance is achieved in extensive experiments, and the results show that the NR performance can be leveraged by the joint training scheme. 
		% \PACS{PACS code1 \and PACS code2 \and more}
		\subclass{Machine Learning \and Computer Vision }
	\end{abstract}
	
	\section{Introduction}
	\label{intro}
	The twenty-first century has witnessed the prospering of the wonderful Internet world. People nowadays not only live in the physical world but also enjoy a virtual life built with information technology. Multimedia documents in various forms like image and video are playing a great role in this online world. Every day there are a huge number of images uploaded or generated on social media and websites like Facebook, Google, Flickr, etc. In pursuit of high-quality life, people tend to seek visual contends of higher quality. In this scenario, being able to predict the perceptual quality of images is becoming significant in a wide range of applications like image compression, video coding, and transmission, image enhancement, image restoration, etc.  
	
	The Image Quality Assessment(IQA) task aims to enable computer system to recognize the perceptual quality level of visual contents. According to the availability of the reference image for the quality prediction of the target image, the IQA task can be divided into three categories: Full-Reference(FR)\cite{bosse2017deep}\cite{ahn2021deep}\cite{shi2021region}\cite{prashnani2018pieapp}
	\cite{ding2020image}\cite{zhang2018unreasonable}\cite{cheon2021perceptual}, Reduced-Reference(RR)\cite{soundararajan2011rred}\cite{wu2016orientation}\cite{zheng2021learning}, and No-Reference methods(NR)\cite{talebi2018nima}
	\cite{wang2021active}\cite{zhu2020metaiqa}\cite{you2021transformer}\cite{ke2021musiq}\cite{zhu2021saliency}\cite{golestaneh2021no}\cite{su2020blindly}\cite{gu2020giqa}. For the FR branch, a variety of methods have been proposed for better alignment to human perception mechanism than naïve Mean Square Error(MSE) on pixel level, among which the Structural Similarity Index(SSIM)\cite{wang2004image} has remained a golden standard in the research community. More over, recent years have witnessed the surge of the wide application of Deep Neural Network on Computer Vision. Early IQA methods often rely on handcrafted features to predict the image quality\cite{zhang2011fsim}\cite{xue2013gradient}\cite{sheikh2006image}. With the help of powerful representation capability of deep convolutional neural network, the sophisticated designing on handcrafted features for quality prediction becomes unnecessary.
	
	However, most of the current deep models only learn local convolutional features inherited from a pre-trained network on recognition tasks like VGG\cite{simonyan2014very} and ResNet\cite{he2016deep}. In real-world scenarios, there can be various distortion types applied to whether locally or globally, within an image. The conventional scheme for deep IQA models with the CNN extractor and MLP score regressor only considers the final representation of the visual content, and attention among different levels is often ignored. 
	
	In this paper, we introduce a hybrid framework consisting of CNN layers and a Transformer encoder. To learn better features for the task of image quality assessment, we build a transformer encoder upon CNN feature extractor. Furthermore, to make the metric surjective, we also add the original pixels as part of the features.  To make the designed model applicable to broad real-world scenarios for image quality assessment, we propose a compatible framework for both FR and NR tasks. To make FR and NR quality prediction compatible in the same framework, we address these two tasks at different levels of the working pipeline. Specifically, inspired by SSIM, we use feature-level statistics to calculate the FR scores between reference and distorted images.
	In contrast, for the NR phase, we address the task as a classification problem instead of directly regressing to a single score. To alleviate the input difference between the two tasks, we employ the Siamese Network structure to extract proper features for quality assessment. Since the model allows for both FR and NR settings, we can optimize the model with different datasets, though some may not have reference images.
	
	The contributions of this paper are summarized as follows:
	
	1.	We introduce an end-to-end deep learning-based method that is compatible with both FR and NR image quality measurement. Previous learning-based methods generally use CNN feature extractor together with a score regressor to figure out the quality score. Since the reference image in the NR setting is missing, the features used for the score regressor are pretty different. This scenario makes it challenging to combine the FR and NR IQA tasks in a unifying framework. To achieve this, we propose to address the FR setting task as a comparison problem on the feature level and the NR setting task as a classification problem. 
	
	2.	We propose a transformer encoder following CNN layers as a complementary feature extractor for the learning process. CNN layers are good at extracting local visual contexts while non-local information is neglected. To address this issue, we utilize the popular transformer model to model the non-local information. 
	
	3.	We carry out image quality experiments for both FR and NR settings on several standard IQA datasets to show the effectiveness of the proposed method.
	
	In the following sections of this paper, we will first give a brief review of the existing models of image quality assessment. Then we introduce our TFIQA method, followed by experiments and results analysis. Finally, we carry out the conclusion.

	\section{Related Work}
	\label{sec:1}
	%Text with citations \cite{RefB} and \cite{RefJ}.
	\subsection{Conventional and Convolutional IQA models}
	\label{sec:2}
	The most simple and well-known metrics for image quality are the MSE distance and the PSNR between the reference image and the distortion image. However, though these metrics are simple and convenient for optimization, they have been proven less correlated to human judgments. By considering more statistical information between the reference and distorted images, SSIM\cite{wang2004image} is proposed as a full reference model, which introduces structure similarity for image quality assessment. It has shown nice alignment to human vision system and inspires many later pieces of research like FSIM\cite{zhang2011fsim} and MS-SSIM\cite{wang2003multiscale}. Another branch of the conventional IQA models are the fidelity based ones such as VSI\cite{zhang2014vsi}, MAD\cite{larson2010most} and VIF\cite{sheikh2006image}.
	
	In recent years, the research community has witnessed great progress made by deep learning. With the help of deep models, now the focus of research on image quality assessment turns from pixel-level or hand-crafted feature level to the automatically learned deep feature level. The LPIPS model introduced by Zhang et al. \cite{zhang2018unreasonable} has shown the effectiveness of applying deep features learned from recognition tasks to the FR-IQA task. Ding et al. propose DISTS\cite{ding2020image} to show even better performance to use structure similarity computing and multi-scale features for FR tasks. DeepIQA\cite{bosse2017deep} employs CNN as a feature extractor and uses two separate fully connected neural networks to predict patch attention weights and patch quality and output the global image quality score with attention-aware pooling on local scores. Su et al. are inspired by the hypernet in vision tasks and propose HyperIQA\cite{su2020blindly} for the NR-IQA task, which can adaptively generate parameters according to the input image for the quality prediction network. Zhu et al. propose MetaIQA\cite{zhu2020metaiqa} to leverage meta-learning to learn IQA models that adapt to different distortion types. 
	
	To overcome the constraints of the small size of datasets and the overfitting issue for the IQA task, a variety of multi-task learning-based models are proposed\cite{kang2015simultaneous}\cite{xu2016multi}\cite{ma2017end}. Another branch of IQA methods to alleviate the limitation of annotated datasets is the ranking based ones\cite{gao2015learning}\cite{ma2017dipiq}\cite{prashnani2018pieapp}\cite{liu2017rankiqa}\cite{ma2019blind}. 
	The ranking-based methods first build the quality ranking datasets(image pairs or lists) and then use the enhanced forms of datasets to train a more complex network for the IQA task. In recent years, the GAN model has been proven effective in generating high-quality images of various resolutions. Some recent works propose to apply the GAN model in IQA task to learn a mapping from distortion image to a hallucinated reference image, guiding the IQA model to learn perceptual differences\cite{lin2018hallucinated} better \cite{pan2018blind}\cite{ren2018ran4iqa}.
	
	\subsection{Transformer based IQA models}
	Transformers\cite{vaswani2017attention} were originally proposed in NLP tasks as a generalized non-local attention mechanism to learn contexts among tokens of arbitrary distance. It has shown excellent performance on language modeling compared to conventional models based on RNN. Apart from language modeling, transformers have shown their effectiveness on vision tasks as well. Vision Transformer has successfully applied a hybrid framework on image recognition,  which consists of a transformer encoder and CNN extractor. 
	Inspired by ViT\cite{dosovitskiy2020image}, TRIQ\cite{you2021transformer} re-define the classification token as the quality embedding token for the NR-IQA task. In order to deal with images of different resolutions, it provides sufficient token numbers to represent the encoded image features. MUSIQ\cite{ke2021musiq} addresses the resolution issue in the IQA task with a different approach, which utilizes patch-based multi-scale mechanism in the transformer encoding process. Similar to that of TRIQ,  the MLP layers are used for the final quality score prediction. In contrast, to deal with the dual input images in the FR setting, the IQT model\cite{cheon2021perceptual} adopts the complete transformer model, which includes both the encoder and decoder. The CNN backbone firstly extracts the deep features of both reference and distorted images. Then the different features are re-encoded by the transformer encoder and input to the decoder as contexts. The deep features of the reference image are also input into the decoder as query information. Finally, in the decoder, the quality embedding token is used for the final quality prediction with the MLP module. Compared to IQT, we propose a more lightweight model for the FR task with only a transformer encoder module. Moreover, instead of learning regression layers for a score prediction, we learn better features and predict quality scores by feature-level statistical comparison.
	
	\subsection{Attention Mechanism for IQA}
	Like the recognition tasks, the spatial and channel attention can also benefit the IQA task\cite{ding2020image}. To assign different impacts caused by different image regions in quality measurements, DeepIQA\cite{bosse2017deep} designs two branches of networks for computing predictions. One branch is for patch importance weights prediction, and the other is for patch score prediction. This kind of attention can be divided to spatial attention. And it is intuitive that different image regions can contribute differently to the overall image quality, especially for the authentic distortion images. The weighting mechanism applied on deep CNN features in the LPIPS\cite{zhang2018unreasonable} and DISTS\cite{ding2020image} shows the contribution made by this attention mechanism on channel levels. As previously mentioned in transformer-based models, the self-attention embedded in the transformer encoder and the cross attention between the encoder and decoder are also proven effective in the IQA task\cite{cheon2021perceptual}.  
	This paper also adopts the compelling deep CNN features as local information and extends with token level attention in the transformer encoding process to get non-local information for proper features predicting final qualities. Considering the different settings of FR and NR tasks, we propose a unified framework by Siamese encoding networks and settle 
	
	%as required. Don't forget to give each section
	%and subsection a unique label (see Sect.~\ref{sec:1}).
	%\paragraph{Paragraph headings} Use paragraph headings as needed.
	%\begin{equation}
	%a^2+b^2=c^2
	%\end{equation}
	
	\section{Methodology}
	\begin{figure}[htbp]
		\centerline{\includegraphics[width=9.5cm]{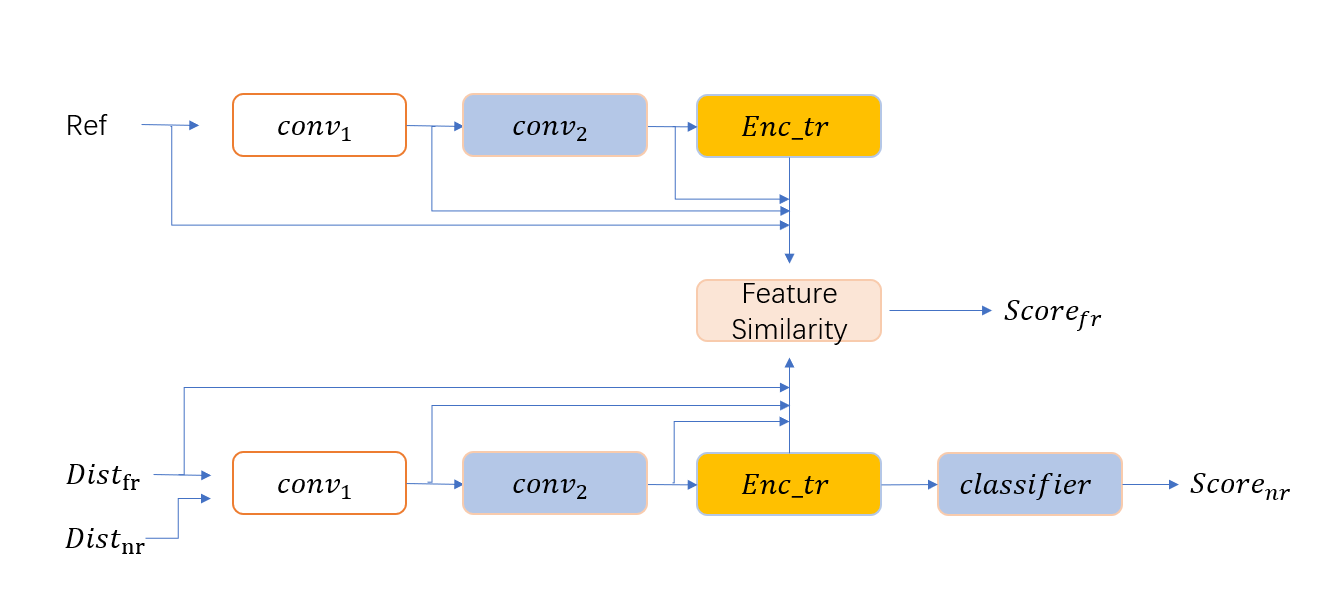}}
		\caption{The overview of the proposed $TFIQA$ framework. Our method utilizes both CNN layers and transformer encoder layers to extract image features. Specifically, our proposed method extends the CNN layers with transformer encoder layers and combines the deep features from both CNN and transformer by attention mechanism, which makes our method different from conventional CNN-based IQA models. The `conv' denotes for convolutional layer, `$Encoder\_tr$' stands for transformer encoder layers.}
		\label{fig:icmc_frame}
	\end{figure}
	
	As shown in Fig. 1, the architecture of our proposed model mainly consists of five modules, including CNN encoder, transformer encoder, attention module, structure similarity for FR score, and NR classifier. The model uses both a full reference dataset that provides reference and distortion image pairs and no reference dataset that only has distortion images. In the training phase, the CNN encoder will first convert the input raw images into deep features, which are the inputs of the shallow transformer encoder. We apply the attention mechanism on both CNN channels and the transformer encoder layers. The feature similarity module will compare the deep features of both reference and distortion images at different channels or layers for the FR branch. The attention module gives adaptive weights on those features maps at different levels. The feature similarity module utilizes SSIM like the structural similarity between feature maps with mean and covariance statistics. We add MLP head as a classifier after the transformer encoder for the NR branch to predict five quality level probabilities.
	
	Our inspiration comes from the trending research on transformers\cite{vaswani2017attention}. CNN is usually employed as the feature extractor and MLP as the score regressor in conventional image quality assessment models. However, CNN is good at merging local contextual information fusion but ignores global and non-local information. Inspired by the success of the transformer in modeling contexts, we propose to use a transformer encoder to refine the features encoded from CNN layers.  Also, from the respective of multi-task learning, the NR and FR tasks are intuitive inter correlated tasks for the shared goal of image quality assessment. Based on this consideration, we re-design the framework of image quality assessment and make it compatible for training both FR and NR branches simultaneously.
	
	Our model aims to learn better deep features for image quality assessment and encourage joint training between the two branches.

	\subsection{CNN Backbone}\label{CNN Encoder}
	
	Herein for the visual features extraction, we use the conventional VGG16\cite{simonyan2014very} as the feature extraction backbone. 
	
	More specifically, following the practice of DISTS\cite{ding2020image}, we use feature maps from five different CNN layers. 
	
	In the second stage of transformer encoding, the CNN features are reshaped as tokens for the transformer model. The information from the feature maps of different layers will be fused by the attention module, especially for the FR branch. During training, the weights of the CNN extractor are fixed.
	
	The feature extracting process can be summarized by the following equations:
	
	\begin{equation}
	{\rm{f}}(i) = VGG(I),
	\end{equation}
	where the notation $f(i)$ stands for the global visual features extracted by the pretrained VGG16 network.

	\subsection{Transformer Encoder}
	In conventional deep image quality assessment models, the image feature extractor mainly consists of CNN layers. In this work, we extend the CNN encoding layers with transformer encoding layers. In order to avoid overfitting, we use a shallow transformer encoder to do the feature learning for this stage. Different from previous work\cite{you2021transformer}\cite{ke2021musiq}\cite{cheon2021perceptual}\cite{zhu2021saliency}, which whether use the transformer encoder only to encode distortion images for NR task or encode the difference features for FR task, we adopt Siamese network structure for the extended encoding process. In other words,  we use a shared transformer encoder for both reference and distortion images to distinguish between images of different visual quality. 
	
	For the structure of the transformer encoder, we follow the conventional configuration of TRIQ\cite{you2021transformer}. Firstly, the input features of shape HxWxC will be converted to NxD by 1x1 convolution as feature projection. Here, H, W, and C stand for the height, width, and channel of the deep CNN features. Moreover, the N is the number of transformer tokens. The D is the dimension of the transformer encoder. For the NR branch, to predict the quality score from the encoded features, we also add the quality token and concatenate it with the image feature tokens. In order to retain the positional information of the tokens, we then add the learnable positional embeddings to the token embeddings.
	
	\begin{figure}[htbp]
		\centerline{\includegraphics[width=5.5cm]{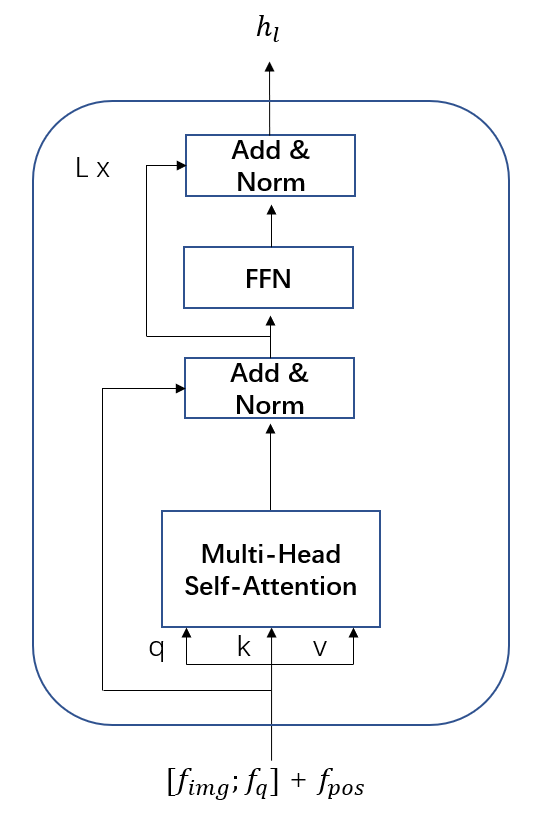}}
		\caption{The inner details of the transformer encoder for feature learning.}
		\label{fig:enc_text}
	\end{figure}
	
	The above encoding process can be summarized as the following equations.
	\begin{equation}
	{y_0} = [{f_q} + {p_q},{f_1} + {p_1},{f_2} + {p_2},...,{f_N} + {p_N}],
	\end{equation}
	where the $g(c)$ represents the final sentence embeddings of a ground truth $c$.
	
	\begin{equation}
	{q_i} \leftarrow {y_{i - 1}},{k_i} \leftarrow {y_{i - 1}},{v_i} \leftarrow {y_{i - 1}},
	\end{equation}
	
	\begin{equation}
	y_i^* = LN(MHA({q_i},{k_i},{v_i}) + {y_{i - 1}}),
	\end{equation}
	
	\begin{equation}
	{y_i} = LN(MLP(y_i^*) + y_i^*),i = 1,..,L
	\end{equation}

	After this finer encoding process for the input images, the model now has collected the relevant features to predict the quality score for FR or NR branches. 
	
	\subsection{Attentive Feature Comparison for FR branch}
	Inspired by LPIPS and DISTS, we proposed an attention module to better predict the quality score for the FR branch. This module is to combine transformer token attention and CNN channel attention for fr image quality assessment. In addition, to make the FR branch model injective, we also add the original image pixels into consideration.
	
	\begin{figure}[htbp]
		\centerline{\includegraphics[width=9.5cm]{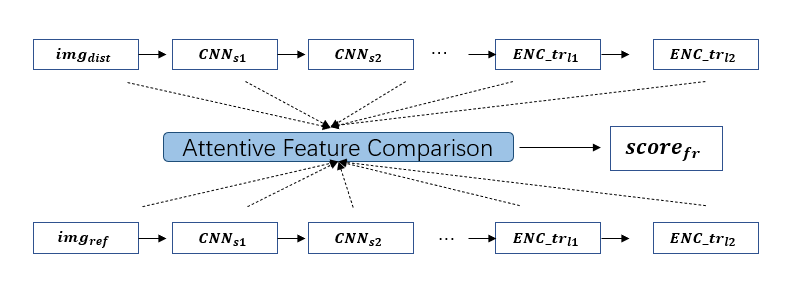}}
		\caption{The inner details of configuration for the FR branch.}
		\label{fig:fr_attention}
	\end{figure}
	
	The attention module is implemented as learnable parameters, which is a constraint as positive values and aims to assign different importance weights to image pixel channels and features at different levels.
	
	As mentioned previously, we extend the widely used CNN layers with transformer encoder. For the feature comparison, we adopt the SSIM like structural statistics of the extracted features, which has been proven effective for FR task in DISTS\cite{ding2020image}.  The modeling process can be expressed with the following equations:
	\begin{equation}
	{{\rm{s}}_\mu }({f_r},{f_d}) = \frac{{2{\mu _{{f_r}}}{\mu _{{f_d}}} + c1}}{{{{({\mu _{{f_r}}})}^2} + {{({\mu _{{f_d}}})}^2} + c1}},
	\end{equation}
	
	\begin{equation}
	{{\rm{s}}_\sigma }({f_r},{f_d}) = \frac{{2{\sigma _{{f_r}{f_d}}} + c2}}{{{{({\sigma _{{f_r}}})}^2} + {{({\sigma _{{f_d}}})}^2} + c2}},
	\end{equation}
	
	\begin{equation}
	Score_{fr}({f_r},{f_d}) =  - \sum\limits_{i = 0}^{i = s} {\sum\limits_{j = 1}^{{c_i}} {(w_{ij}^\mu {s_\mu }(f_{{r_j}}^i,f_{{d_j}}^i) + w_{ij}^\sigma {s_\sigma }(f_{{r_j}}^i,f_{{d_j}}^i))} }  + 1,
	\end{equation}
	where $w$ are the different weights applied on the difference terms of different feature levels. The index $i$ indicates the stage of features, including image pixels as stage zero, five convolutional stages and transformer encoding layers. And for each feature stage, there are different number of channels or tokens, which is indicated by the index $j$. The upper scripts of $\mu$ and $\sigma$ are labels to indicate whether the weights applied on $\rm{s}_\sigma$ or $\rm{s}_\mu$. The $f_r$ and $f_d$ are the feature vectors of reference and distortion images at certain levels indicated by $i$ and $j$, respectively. In order to achieve the positive weights, we need to clip the values of attention weights before gradient back-propagation.
	
	Our general goal is to utilize the transformer encoder to get better visual representations for the task of image quality assessment. And in our overall framework, we share the visual encoder for both reference and distortion images. In the next section, we take a step further and share the visual encoding framework of our proposed method to the NR task. In other words, we add the NR branch beside the original FR branch.
	
	\subsection{Classification for NR quality assessment}
	For the NR branch, the quality prediction is calculated by the MLP head once the visual features are collected. The prediction head consists of two linear layers with the relu activation in between. The prediction head receives the learned quality embeddings from the transformer encoder. 
	
	\begin{figure}[htbp]
		\centerline{\includegraphics[width=5.5cm]{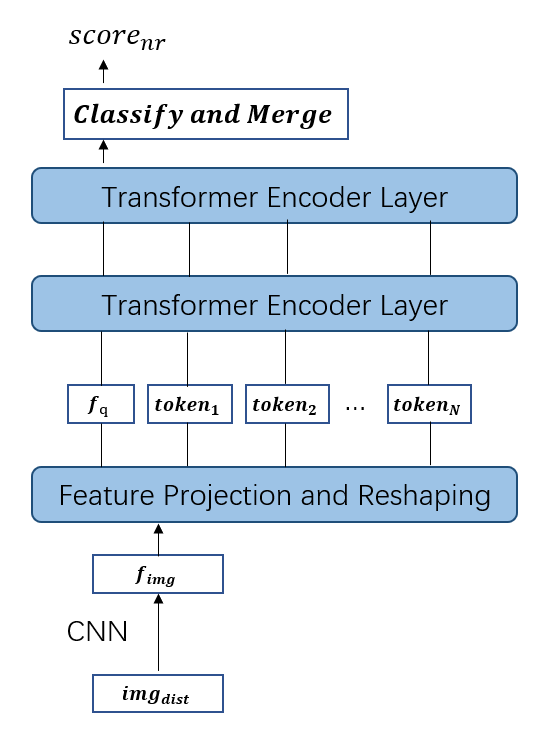}}
		\caption{Structure Details for NR branch.}
		\label{fig:nr_regression}
	\end{figure}
	
	In the Classifier, the first layer will project the features of the transformer hidden dimension into the MLP head dimension. Then the activation layer is introduced for non-linearity. Finally, the second linear layer will map the hidden feature into a distribution over five classes as conducted in NIMA model\cite{talebi2018nima}. Finally, the predicted probability distribution will be merged into a quality score, which is denoted as $Score_{nr}$.
	
	\subsection{Model Training}
	On image quality learning the commonly used loss function is the MAE and MSE loss between the output of IQA model and the ground truth scores. It is expressed as the following equation:
	\begin{equation}
	{L_{fr}} = MSE(Scor{e_{fr}},labe{l_{fr}}),
	\end{equation}
	
	\begin{equation}
	{L_{nr}} = EMD({p_{nr}},labe{l_{nr}}),
	\end{equation}
	
	\begin{equation}
	{L_{all}} = {L_{fr}} + {L_{nr}},
	\end{equation}
	
	In this work, Considering the difference of FR and NR tasks, the score prediction for the FR branch is calculated by feature level comparison, while a classifier is placed over the transformer layers to learn to classify on the quality levels. We use the MSE loss to guide the learning process of the FR branch, which could be expressed with the equation (9). The NR branch is guided by a classification loss, as shown in equation (10).
	
	Above all, the proposed framework is compatible with training with both FR and NR settings. The two branches can both be learned separately and can also be trained in a joined manner, which will be discussed in detail in the next section.

	\section{Experiments}
	\subsection{Dataset and Evaluation Metrics}
	
	\textbf{Dataset.} We evaluate the proposed IQA model on several publicly available IQA datasets, including synthetic and authentic types. For training the NR branch, we use the KONIQ-10K dataset\cite{hosu2020koniq}. For this dataset, we randomly split eighty percent and twenty percent for training and testing, respectively. For training the FR branch, we use the KADID-10K dataset\cite{lin2019kadid}. The performance evaluation for the FR branch is conducted on three standard IQA datasets, i.e., LIVE\cite{sheikh2006statistical}, CSIQ\cite{larson2010most} and TID2013\cite{ponomarenko2013color}.
	
	\textbf{Metrics.}  For performance evaluation, we use both correlation index and RMSE differences between the predicted scores and MOS scores.
	
	\textbf{Baseline and SOTA models.} We compare our proposed method to both conventional and CNN-based ones. In addition, the recent transformer-based model IQT is also included. For simple notation, we just name them LPIPS, DISTS, DeepIQA, respectively. And our proposed transformer-based method is noted as TFIQA. The IQT model utilizes both transformer encoder and decoder for FR image quality assessment. The LPIPS and DISTS model are CNN-based FR IQA models. 
	
	\subsection{Implementing Details}
	In this paper, we propose a framework to enhance the CNN features with a sequential transformer encoder for the image quality assessment task. This framework is compatible with both full reference and no reference configurations. For the full reference mode, we denote the corresponding model as $\rm{TFIQA\_{UNI}}$. For this model, the VGG16 is used as the CNN backbone feature extractor, and the hyper-parameters of the transformer encoder are set following the configuration of TRIQ\cite{you2021transformer}. Firstly, we choose a shallow transformer encoder and set the encoding layers as 2. For the multi-head self-attention, we set the number of heads as 4. And the transformer encoder dimension is set as 256. The hidden dimensions of MLP both in transformer encoder layers and the prediction head are set as 1024. For the  MLP prediction head, we add dropout as 0.1. The initial learning rate is set as 2e-4. And the Adam optimizer is utilized for learning the model parameters.
	
	In addition, we also consider the no-reference scenario for the quality assessment task. Since in this mode, we only have the distortion image to predict the quality score, the feature comparison method adopted in the full reference mode is not applicable in this phase. And comparing to directly predicting a score from image features, we think it is more intuitive to learn a classifier for the prediction. And following the work of NIMA\cite{talebi2018nima}, we utilize a five classes predictor and use the probability values to calculate the final quality score.
	
	For the no reference mode, we denote our model as $\rm{TFIQA\_{UNI}}$. For this mode, the data augmentation strategy is used, i.e., we use image patches for training instead of the whole image so that the model can have enough training data to converge. And to overcome the possibility of overfitting, a smaller configuration size is utilized. Again we set the transformer encoding layers as 2. The model dimension of the transformer encoder is set as 32, and the number of heads for multi-head self-attention is set as 8. The hidden dimensions of the MLP layer in both the prediction head and the encoding layers are set as 64. And the prediction head the number of quality levels is set as 5. In the MLP layers, a dropout rate of 0.1 is used. The initial learning rate is set as 0.5e-4, and the SGD optimizer is configured with a momentum of 0.9.

	\subsection{Experiments Results and Analysis}
	\textbf{Comparisons with state-of-the-arts on FR performance.}
	In this section, we conduct experiments to evaluate the full reference performance of the proposed method $\rm{TFIQA\_{FR}}$ comparing with both the contemporary metrics and learning-based models. Note that our $\rm{TFIQA\_{FR}}$ model is trained on KADID-10K dataset. And some of the results are borrowed from \cite{cheon2021perceptual} and \cite{ding2020image}.  
	
	From the above results, we can see that our model overperforms the conventional full reference models like PSNR and CNN-based models like DeepIQA and PieAPP. This should validate the effectiveness of our proposed method on the full reference scenario. The performance of our proposed model is better than LPIPS and DISTS nearly on every metric and evaluation dataset. We think this is mainly due to the differences in the feature learning process. The LPIPS model and DISTS model only have CNN layers to extract the image features for quality prediction while the non-local information is ignored. With the proposed $\rm{TFIQA\_{FR}}$ model, we extend the CNN layers with transformer encoders, which employ the multi-head self-attention to model the contextual information in the images. Also, following the practice of DISTS, we apply attention weights to the deep features in different channels.  Different from the DISTS model, which only has CNN features to attend to, we attend to both CNN features and transformer features. For the transformer features, we also observe that it is better to apply weights in the form of token embeddings instead of reforming the token embeddings into two-dimensional feature maps like the form of deep convolutional feature maps.
	
	For the recently proposed transformer-based IQT model\cite{cheon2021perceptual}, the performance of our method is also comparable. Note that the IQT mode utilizes both transformer encoder and decoder for the modeling, while our method only employs the encoder layers. And the way to use the image features is also different. We use a statistical comparison on deep features to compute the quality score, while IQT uses regression to predict the score. Note that we utilize a Siamese structure for the image encoding process, which means both the reference and the distortion images are processed with the same visual encoder. It is feasible to unify the full reference and no reference prediction mode in the same model framework based on this setting.
	
	\begin{table*}[]
		\centering
		\caption{Performance evaluations on the three standard IQA Dataset in terms of PLCC and SROCC. tfiqa-fr is our proposed full reference model.}  % 
		\label{fr performance}       % Give a unique label
		\begin{tabular}{lllllll}
			\hline
			& LIVE   &        & CSIQ   &        & TID2013 &        \\ \cline{2-7} 
			& PLCC   & SROCC  & PLCC   & SROCC  & PLCC    & SROCC  \\ \hline
			PSNR     & 0.865  & 0.873  & 0.819  & 0.810   & 0.677   & 0.687  \\
			SSIM\cite{wang2004image}     & 0.937  & 0.948  & 0.852  & 0.865  & 0.777   & 0.727  \\
			MS-SSIM\cite{wang2003multiscale}  & 0.940   & 0.951  & 0.889  & 0.906  & 0.83    & 0.786  \\
			VSI\cite{zhang2014vsi}      & 0.948  & 0.952  & 0.928  & 0.942  & \textbf{0.900}     & \textbf{0.897}  \\
			MAD\cite{larson2010most}      & \textbf{0.968}  & \textbf{0.967}  & \textbf{0.950}   & 0.947  & 0.827   & 0.781  \\
			VIF\cite{sheikh2006image}      & 0.960   & 0.964  & 0.913  & 0.911  & 0.771   & 0.677  \\
			FSIM\cite{zhang2011fsim}     & 0.961  & 0.965  & 0.919  & 0.931  & 0.877   & 0.851  \\
			NLPD\cite{laparra2016perceptual}     & 0.932  & 0.937  & 0.923  & 0.932  & 0.839   & 0.800    \\
			GMSD\cite{xue2013gradient}     & 0.957  & 0.960   & 0.945  & \textbf{0.950}   & 0.855   & 0.804  \\
			\hline
			DeepIQA\cite{bosse2017deep}  & 0.940   & 0.947  & 0.901  & 0.909  & 0.834   & 0.831  \\
			PieAPP\cite{prashnani2018pieapp}   & 0.908  & 0.919  & 0.877  & 0.892  & 0.859   & 0.876  \\
			LPIPS\cite{zhang2018unreasonable}    & 0.934  & 0.932  & 0.896  & 0.876  & 0.749   & 0.670   \\
			DISTS\cite{ding2020image}    & \textbf{0.954}  & 0.954  & 0.928  & 0.929  & 0.855   & 0.830   \\
			IQT\cite{cheon2021perceptual}      & *      & \textbf{0.970}   & *      & \textbf{0.943}  & *       & \textbf{0.899}  \\
			$\rm{TFIQA\_{FR}}$ & 0.947 & 0.958 & \textbf{0.941} & 0.938 & \textbf{0.858}  & 0.832 \\ \hline
		\end{tabular}
	\end{table*}

	\textbf{Learning with NR branch.} Most of the deep models form the problem of no-reference image quality assessment as a regression problem. However, compared to the large scale of datasets in conventional deep learning tasks like image classification, there are relatively limited annotations provided for the task of image quality assessment. In this scenario, we apply the patch-based training strategy for NR quality assessment following \cite{su2020blindly}. For our model configured as NR mode, we also train with this patch strategy. In addition, we adopt a classifier and form the score prediction as a classification problem following the practice of \cite{talebi2018nima}. The results are shown in Table 2. As we can see, our $\rm{TFIQA\_{NR}}$ model achieves better performance than the CNN-based model $\rm{DeepIQA\_{NR}}$. However, comparing to the MetaIQA model\cite{zhu2020metaiqa}, there is still a performance gap for the $\rm{TFIQA\_{NR}}$ model. We think this should own to the gaining from multi-task training of the MetaIQA model.
	
	\begin{table*}[]
		\centering
		\caption{Non-reference Performance evaluations on KONIQ-10K in terms of PLCC, SROCC and RMSE.}  % 
		\label{tab:nr performance}       % Give a unique label
		\begin{tabular}{llll}
			\hline
			& PLCC   & SROCC  & RMSE   \\ \hline
			$\rm{DeepIQA\_{NR}}$\cite{bosse2017deep} & 0.761  & 0.739  & *      \\
			MetaIQA\cite{zhu2020metaiqa}    & \textbf{0.887}  & \textbf{0.850}   & *      \\
			$\rm{TFIQA\_{NR}}$    & 0.808 & 0.769 & 0.349 \\
			$\rm{TFIQA\_{UNI}}$   & \textbf{0.853} & \textbf{0.836} & \textbf{0.291} \\ \hline
		\end{tabular}
	\end{table*}
	\textbf{Improving the NR performance with FR branch.} Previously, we have shown that the proposed method can be applied to both FR and NR mode. From previous results, we also observe that comparing the FR mode, the performance of  NR mode has larger space to improve. Since our proposed framework is compatible for both FR and NR mode, in this section we consider whether the NR branch can be improved by a joint training scheme together with FR branch. Comparing to the original triq-nr model, we add another FR branch to allow input of the reference images. And the new model is denoted as triq-uni. Note that for both branches, the visual encoder is shared as a Siemese structure. For the training of triq-uni, we forward the two branches with different datasets. The NR branch is forwarded with KONIQ-10K dataset and the FR branch with KADID-10K dataset. The performances are shown in Table 2. We can see that comparing the previous model $\rm{TFIQA\_{NR}}$, the $\rm{TFIQA\_{UNI}}$ achieves better performance in every evaluation metric, which should validate the effectiveness of joint training scheme over the proposed unified framework.

	\section{Conclusion}
	In this paper, we discuss a unified transformer-based framework that is compatible with both NR and FR image quality assessment tasks. Specifically, we extend the conventional CNN feature extractor with a transformer encoder. For the FR configuration, we utilize channel attention to assign different weights on deep CNN features from different channels and Transformer features from different tokens. The final quality scores are predicted by structural statistical comparison of the deep features with attention weights. The comparisons with state-of-the-art IQA models are also provided. And the results show that our proposed FR model achieves outstanding performance among all metrics.
	
	In addition, we also explore the NR configuration of the proposed method. Several training strategies are considered, like patch-based training, data augmentation, and an FR-aided training scheme to gain better prediction accuracy. Experiment results show that our proposed NR method can achieve better performance than the conventional CNN-based model. And the results also show that a joint training scheme together with the FR branch can achieve good performance gains. Future work may relate to better generalization on various distortion types and introduce a better feature learning scheme for the image quality assessment task.

	\section*{Acknowledgment}
	
	This work was supported by grants from the Research Grants Council of the Hong Kong Special Administrative Region, China, and from the City University of Hong Kong.

	\bibliographystyle{IEEEtran}
	\bibliography{refs.bib}

% Generated by IEEEtran.bst, version: 1.14 (2015/08/26)
\begin{thebibliography}{10}
\providecommand{\url}[1]{#1}
\csname url@samestyle\endcsname
\providecommand{\newblock}{\relax}
\providecommand{\bibinfo}[2]{#2}
\providecommand{\BIBentrySTDinterwordspacing}{\spaceskip=0pt\relax}
\providecommand{\BIBentryALTinterwordstretchfactor}{4}
\providecommand{\BIBentryALTinterwordspacing}{\spaceskip=\fontdimen2\font plus
\BIBentryALTinterwordstretchfactor\fontdimen3\font minus
  \fontdimen4\font\relax}
\providecommand{\BIBforeignlanguage}[2]{{%
\expandafter\ifx\csname l@#1\endcsname\relax
\typeout{** WARNING: IEEEtran.bst: No hyphenation pattern has been}%
\typeout{** loaded for the language `#1'. Using the pattern for}%
\typeout{** the default language instead.}%
\else
\language=\csname l@#1\endcsname
\fi
#2}}
\providecommand{\BIBdecl}{\relax}
\BIBdecl

\bibitem{bosse2017deep}
S.~Bosse, D.~Maniry, K.-R. M{\"u}ller, T.~Wiegand, and W.~Samek, ``Deep neural
  networks for no-reference and full-reference image quality assessment,''
  \emph{IEEE Transactions on image processing}, vol.~27, no.~1, pp. 206--219,
  2017.

\bibitem{ahn2021deep}
S.~Ahn, Y.~Choi, and K.~Yoon, ``Deep learning-based distortion sensitivity
  prediction for full-reference image quality assessment,'' in
  \emph{Proceedings of the IEEE/CVF Conference on Computer Vision and Pattern
  Recognition}, 2021, pp. 344--353.

\bibitem{shi2021region}
S.~Shi, Q.~Bai, M.~Cao, W.~Xia, J.~Wang, Y.~Chen, and Y.~Yang,
  ``Region-adaptive deformable network for image quality assessment,'' in
  \emph{Proceedings of the IEEE/CVF Conference on Computer Vision and Pattern
  Recognition}, 2021, pp. 324--333.

\bibitem{prashnani2018pieapp}
E.~Prashnani, H.~Cai, Y.~Mostofi, and P.~Sen, ``Pieapp: Perceptual image-error
  assessment through pairwise preference,'' in \emph{Proceedings of the IEEE
  Conference on Computer Vision and Pattern Recognition}, 2018, pp. 1808--1817.

\bibitem{ding2020image}
K.~Ding, K.~Ma, S.~Wang, and E.~P. Simoncelli, ``Image quality assessment:
  Unifying structure and texture similarity,'' \emph{arXiv preprint
  arXiv:2004.07728}, 2020.

\bibitem{zhang2018unreasonable}
R.~Zhang, P.~Isola, A.~A. Efros, E.~Shechtman, and O.~Wang, ``The unreasonable
  effectiveness of deep features as a perceptual metric,'' in \emph{Proceedings
  of the IEEE conference on computer vision and pattern recognition}, 2018, pp.
  586--595.

\bibitem{cheon2021perceptual}
M.~Cheon, S.-J. Yoon, B.~Kang, and J.~Lee, ``Perceptual image quality
  assessment with transformers,'' in \emph{Proceedings of the IEEE/CVF
  Conference on Computer Vision and Pattern Recognition}, 2021, pp. 433--442.

\bibitem{soundararajan2011rred}
R.~Soundararajan and A.~C. Bovik, ``Rred indices: Reduced reference entropic
  differencing for image quality assessment,'' \emph{IEEE Transactions on Image
  Processing}, vol.~21, no.~2, pp. 517--526, 2011.

\bibitem{wu2016orientation}
J.~Wu, W.~Lin, G.~Shi, L.~Li, and Y.~Fang, ``Orientation selectivity based
  visual pattern for reduced-reference image quality assessment,''
  \emph{Information Sciences}, vol. 351, pp. 18--29, 2016.

\bibitem{zheng2021learning}
H.~Zheng, H.~Yang, J.~Fu, Z.-J. Zha, and J.~Luo, ``Learning conditional
  knowledge distillation for degraded-reference image quality assessment,'' in
  \emph{Proceedings of the IEEE/CVF International Conference on Computer
  Vision}, 2021, pp. 10\,242--10\,251.

\bibitem{talebi2018nima}
H.~Talebi and P.~Milanfar, ``Nima: Neural image assessment,'' \emph{IEEE
  Transactions on Image Processing}, vol.~27, no.~8, pp. 3998--4011, 2018.

\bibitem{wang2021active}
Z.~Wang and K.~Ma, ``Active fine-tuning from gmad examples improves blind image
  quality assessment,'' \emph{IEEE Transactions on Pattern Analysis and Machine
  Intelligence}, 2021.

\bibitem{zhu2020metaiqa}
H.~Zhu, L.~Li, J.~Wu, W.~Dong, and G.~Shi, ``Metaiqa: Deep meta-learning for
  no-reference image quality assessment,'' in \emph{Proceedings of the IEEE/CVF
  Conference on Computer Vision and Pattern Recognition}, 2020, pp.
  14\,143--14\,152.

\bibitem{you2021transformer}
J.~You and J.~Korhonen, ``Transformer for image quality assessment,'' in
  \emph{2021 IEEE International Conference on Image Processing (ICIP)}.\hskip
  1em plus 0.5em minus 0.4em\relax IEEE, 2021, pp. 1389--1393.

\bibitem{ke2021musiq}
J.~Ke, Q.~Wang, Y.~Wang, P.~Milanfar, and F.~Yang, ``Musiq: Multi-scale image
  quality transformer,'' in \emph{Proceedings of the IEEE/CVF International
  Conference on Computer Vision}, 2021, pp. 5148--5157.

\bibitem{zhu2021saliency}
M.~Zhu, G.~Hou, X.~Chen, J.~Xie, H.~Lu, and J.~Che, ``Saliency-guided
  transformer network combined with local embedding for no-reference image
  quality assessment,'' in \emph{Proceedings of the IEEE/CVF International
  Conference on Computer Vision}, 2021, pp. 1953--1962.

\bibitem{golestaneh2021no}
S.~A. Golestaneh, S.~Dadsetan, and K.~M. Kitani, ``No-reference image quality
  assessment via transformers, relative ranking, and self-consistency,''
  \emph{arXiv preprint arXiv:2108.06858}, 2021.

\bibitem{su2020blindly}
S.~Su, Q.~Yan, Y.~Zhu, C.~Zhang, X.~Ge, J.~Sun, and Y.~Zhang, ``Blindly assess
  image quality in the wild guided by a self-adaptive hyper network,'' in
  \emph{Proceedings of the IEEE/CVF Conference on Computer Vision and Pattern
  Recognition}, 2020, pp. 3667--3676.

\bibitem{gu2020giqa}
S.~Gu, J.~Bao, D.~Chen, and F.~Wen, ``Giqa: Generated image quality
  assessment,'' in \emph{European Conference on Computer Vision}.\hskip 1em
  plus 0.5em minus 0.4em\relax Springer, 2020, pp. 369--385.

\bibitem{wang2004image}
Z.~Wang, A.~C. Bovik, H.~R. Sheikh, and E.~P. Simoncelli, ``Image quality
  assessment: from error visibility to structural similarity,'' \emph{IEEE
  transactions on image processing}, vol.~13, no.~4, pp. 600--612, 2004.

\bibitem{zhang2011fsim}
L.~Zhang, L.~Zhang, X.~Mou, and D.~Zhang, ``Fsim: A feature similarity index
  for image quality assessment,'' \emph{IEEE transactions on Image Processing},
  vol.~20, no.~8, pp. 2378--2386, 2011.

\bibitem{xue2013gradient}
W.~Xue, L.~Zhang, X.~Mou, and A.~C. Bovik, ``Gradient magnitude similarity
  deviation: A highly efficient perceptual image quality index,'' \emph{IEEE
  Transactions on Image Processing}, vol.~23, no.~2, pp. 684--695, 2013.

\bibitem{sheikh2006image}
H.~R. Sheikh and A.~C. Bovik, ``Image information and visual quality,''
  \emph{IEEE Transactions on image processing}, vol.~15, no.~2, pp. 430--444,
  2006.

\bibitem{simonyan2014very}
K.~Simonyan and A.~Zisserman, ``Very deep convolutional networks for
  large-scale image recognition,'' \emph{arXiv preprint arXiv:1409.1556}, 2014.

\bibitem{he2016deep}
K.~He, X.~Zhang, S.~Ren, and J.~Sun, ``Deep residual learning for image
  recognition,'' in \emph{Proceedings of the IEEE conference on computer vision
  and pattern recognition}, 2016, pp. 770--778.

\bibitem{wang2003multiscale}
Z.~Wang, E.~P. Simoncelli, and A.~C. Bovik, ``Multiscale structural similarity
  for image quality assessment,'' in \emph{The Thrity-Seventh Asilomar
  Conference on Signals, Systems \& Computers, 2003}, vol.~2.\hskip 1em plus
  0.5em minus 0.4em\relax Ieee, 2003, pp. 1398--1402.

\bibitem{zhang2014vsi}
L.~Zhang, Y.~Shen, and H.~Li, ``Vsi: A visual saliency-induced index for
  perceptual image quality assessment,'' \emph{IEEE Transactions on Image
  processing}, vol.~23, no.~10, pp. 4270--4281, 2014.

\bibitem{larson2010most}
E.~C. Larson and D.~M. Chandler, ``Most apparent distortion: full-reference
  image quality assessment and the role of strategy,'' \emph{Journal of
  electronic imaging}, vol.~19, no.~1, p. 011006, 2010.

\bibitem{kang2015simultaneous}
L.~Kang, P.~Ye, Y.~Li, and D.~Doermann, ``Simultaneous estimation of image
  quality and distortion via multi-task convolutional neural networks,'' in
  \emph{2015 IEEE international conference on image processing (ICIP)}.\hskip
  1em plus 0.5em minus 0.4em\relax IEEE, 2015, pp. 2791--2795.

\bibitem{xu2016multi}
L.~Xu, J.~Li, W.~Lin, Y.~Zhang, L.~Ma, Y.~Fang, and Y.~Yan, ``Multi-task rank
  learning for image quality assessment,'' \emph{IEEE Transactions on Circuits
  and Systems for Video Technology}, vol.~27, no.~9, pp. 1833--1843, 2016.

\bibitem{ma2017end}
K.~Ma, W.~Liu, K.~Zhang, Z.~Duanmu, Z.~Wang, and W.~Zuo, ``End-to-end blind
  image quality assessment using deep neural networks,'' \emph{IEEE
  Transactions on Image Processing}, vol.~27, no.~3, pp. 1202--1213, 2017.

\bibitem{gao2015learning}
F.~Gao, D.~Tao, X.~Gao, and X.~Li, ``Learning to rank for blind image quality
  assessment,'' \emph{IEEE transactions on neural networks and learning
  systems}, vol.~26, no.~10, pp. 2275--2290, 2015.

\bibitem{ma2017dipiq}
K.~Ma, W.~Liu, T.~Liu, Z.~Wang, and D.~Tao, ``dipiq: Blind image quality
  assessment by learning-to-rank discriminable image pairs,'' \emph{IEEE
  Transactions on Image Processing}, vol.~26, no.~8, pp. 3951--3964, 2017.

\bibitem{liu2017rankiqa}
X.~Liu, J.~Van De~Weijer, and A.~D. Bagdanov, ``Rankiqa: Learning from rankings
  for no-reference image quality assessment,'' in \emph{Proceedings of the IEEE
  International Conference on Computer Vision}, 2017, pp. 1040--1049.

\bibitem{ma2019blind}
K.~Ma, X.~Liu, Y.~Fang, and E.~P. Simoncelli, ``Blind image quality assessment
  by learning from multiple annotators,'' in \emph{2019 IEEE International
  Conference on Image Processing (ICIP)}.\hskip 1em plus 0.5em minus
  0.4em\relax IEEE, 2019, pp. 2344--2348.

\bibitem{lin2018hallucinated}
K.-Y. Lin and G.~Wang, ``Hallucinated-iqa: No-reference image quality
  assessment via adversarial learning,'' in \emph{Proceedings of the IEEE
  Conference on Computer Vision and Pattern Recognition}, 2018, pp. 732--741.

\bibitem{pan2018blind}
D.~Pan, P.~Shi, M.~Hou, Z.~Ying, S.~Fu, and Y.~Zhang, ``Blind predicting
  similar quality map for image quality assessment,'' in \emph{Proceedings of
  the IEEE conference on computer vision and pattern recognition}, 2018, pp.
  6373--6382.

\bibitem{ren2018ran4iqa}
H.~Ren, D.~Chen, and Y.~Wang, ``Ran4iqa: Restorative adversarial nets for
  no-reference image quality assessment,'' in \emph{Thirty-Second AAAI
  Conference on Artificial Intelligence}, 2018.

\bibitem{vaswani2017attention}
A.~Vaswani, N.~Shazeer, N.~Parmar, J.~Uszkoreit, L.~Jones, A.~N. Gomez,
  {\L}.~Kaiser, and I.~Polosukhin, ``Attention is all you need,'' in
  \emph{Advances in neural information processing systems}, 2017, pp.
  5998--6008.

\bibitem{dosovitskiy2020image}
A.~Dosovitskiy, L.~Beyer, A.~Kolesnikov, D.~Weissenborn, X.~Zhai,
  T.~Unterthiner, M.~Dehghani, M.~Minderer, G.~Heigold, S.~Gelly \emph{et~al.},
  ``An image is worth 16x16 words: Transformers for image recognition at
  scale,'' \emph{arXiv preprint arXiv:2010.11929}, 2020.

\bibitem{hosu2020koniq}
V.~Hosu, H.~Lin, T.~Sziranyi, and D.~Saupe, ``Koniq-10k: An ecologically valid
  database for deep learning of blind image quality assessment,'' \emph{IEEE
  Transactions on Image Processing}, vol.~29, pp. 4041--4056, 2020.

\bibitem{lin2019kadid}
H.~Lin, V.~Hosu, and D.~Saupe, ``Kadid-10k: A large-scale artificially
  distorted iqa database,'' in \emph{2019 Eleventh International Conference on
  Quality of Multimedia Experience (QoMEX)}.\hskip 1em plus 0.5em minus
  0.4em\relax IEEE, 2019, pp. 1--3.

\bibitem{sheikh2006statistical}
H.~R. Sheikh, M.~F. Sabir, and A.~C. Bovik, ``A statistical evaluation of
  recent full reference image quality assessment algorithms,'' \emph{IEEE
  Transactions on image processing}, vol.~15, no.~11, pp. 3440--3451, 2006.

\bibitem{ponomarenko2013color}
N.~Ponomarenko, O.~Ieremeiev, V.~Lukin, K.~Egiazarian, L.~Jin, J.~Astola,
  B.~Vozel, K.~Chehdi, M.~Carli, F.~Battisti \emph{et~al.}, ``Color image
  database tid2013: Peculiarities and preliminary results,'' in \emph{european
  workshop on visual information processing (EUVIP)}.\hskip 1em plus 0.5em
  minus 0.4em\relax IEEE, 2013, pp. 106--111.

\bibitem{laparra2016perceptual}
V.~Laparra, J.~Ball{\'e}, A.~Berardino, and E.~P. Simoncelli, ``Perceptual
  image quality assessment using a normalized laplacian pyramid,''
  \emph{Electronic Imaging}, vol. 2016, no.~16, pp. 1--6, 2016.

\end{thebibliography}
\end{document}